\documentclass{article}
\usepackage{iclr2027,times}

\usepackage{amsmath,amsfonts,bm}
\usepackage{caption}
\def\eqref#1{equation~\ref{#1}}
\def\1{\bm{1}}

\def\rmK{{\mathbf{K}}}

\def\rmT{{\mathbf{T}}}

\def\rmV{{\mathbf{V}}}

\DeclareMathAlphabet{\mathsfit}{\encodingdefault}{\sfdefault}{m}{sl}
\SetMathAlphabet{\mathsfit}{bold}{\encodingdefault}{\sfdefault}{bx}{n}

\usepackage{amsmath}
\usepackage{amssymb}
\usepackage{array}
\usepackage{booktabs}
\usepackage{graphicx}
\usepackage{wrapfig}
\usepackage{multirow}
\usepackage{makecell}
\usepackage{enumitem}
\usepackage{tikz}
\usepackage{colortbl}
\usetikzlibrary{arrows.meta,positioning}
\usepackage{hyperref}
\hypersetup{hidelinks}
\usepackage{url}

\setlist[itemize]{leftmargin=*, itemsep=2pt, topsep=3pt}

\newcommand{\sgdetpp}{SGDet3D++}
\newcommand{\method}{\sgdetpp}
\newcommand{\sgdet}{SGDet3D}

\definecolor{best}{RGB}{255,180,180}
\definecolor{second}{RGB}{255,230,180}
\definecolor{third}{RGB}{255,255,200}
\definecolor{besterror}{RGB}{210,230,255}
\definecolor{seconderror}{RGB}{228,240,255}
\definecolor{thirderror}{RGB}{240,247,255}
\newcommand{\bestcell}[1]{\cellcolor{best}\textbf{#1}}
\newcommand{\secondcell}[1]{\cellcolor{second}#1}
\newcommand{\thirdcell}[1]{\cellcolor{third}#1}
\newcommand{\besterrcell}[1]{\cellcolor{besterror}\textbf{#1}}
\newcommand{\seconderrcell}[1]{\cellcolor{seconderror}#1}
\newcommand{\thirderrcell}[1]{\cellcolor{thirderror}#1}
\newcommand{\na}{--}
\newcommand{\abresult}[2]{#1{\footnotesize$\,\pm\,$#2}}
\newcolumntype{C}[1]{>{\centering\arraybackslash}p{#1}}
\newcolumntype{L}[1]{>{\raggedright\arraybackslash}p{#1}}

\title{\sgdetpp{}: Geometry-Grounded Semantics for 4D Radar--Camera 3D Object Detection}

\author{
Xiaokai Bai\textsuperscript{1},
Zhenyu Fan\textsuperscript{1},
Lianqing Zheng\textsuperscript{2},
Songkai Wang\textsuperscript{1},
Siyuan Cao\textsuperscript{1},
Hui-liang Shen\textsuperscript{1}
\\[2mm]
\textsuperscript{1}College of Information Science and Electronic Engineering, Zhejiang University
\\
\textsuperscript{2}School of Automotive Studies, Tongji University
\\
\texttt{email:shawnnnkb@gmail.com}
}
\iclrfinalcopy

\begin{document}

\maketitle

\begin{abstract}
    4D radar complements dense image semantics with long-range geometry and
    radial motion, but existing radar--camera detectors largely solve
    \emph{where} to align the modalities while leaving \emph{whether} a piece
    of evidence supports an evolving object hypothesis implicit. An image token
    may describe an occluder, a nearby radar return may belong to another object,
    and a pose-aligned memory slot may carry incompatible motion. We formulate
    \emph{hypothesis-conditioned evidence grounding}, which separates candidate
    access from evidence use: semantic, geometric, or temporal evidence is
    filtered or conditioned by the evolving 3D state before updating the
    corresponding query. \sgdetpp{} instantiates this principle through
    Anchor-Grounded Semantic Retrieval (AGR), which conditions deformable image
    retrieval on pooled anchor-consistent radar support; Geometry-Consistent
    Anchor Refinement (GCR), which attentively aggregates individual associated
    returns; and Doppler-Verified Correspondence (DVC), which replaces history
    only when current radial motion contradicts it. \sgdetpp{} improves the
    strongest compared method by 3.82 mAP and 6.82 ODS on OmniHD-Scenes and by
    6.82 mAP and 9.22 NDS on ManTruckScenes, while also leading the listed
    methods in the TJ4DRadSet test comparison. Mechanism-targeted evaluations
    show that AGR improves strict AP
    in every projected-occlusion bin, the yaw-aligned box gate raises
    target-return purity from 29.95\% to 58.87\%, and DVC preserves 96.11\%
    of motion-consistent history
    while retaining 75.90\% contradiction recall. Code will be released.

    % This ambiguity makes it difficult for each object query to determine
    % which semantic, geometric, and historical evidence should be trusted.
\end{abstract}
\vspace{-10pt}
\section{Introduction}
\label{sec:introduction}

Reliable 3D object detection must localize dynamic objects under changing
visibility, range, and traffic density. Surround-view cameras provide dense
appearance and category cues, but remain ambiguous along the viewing ray. 4D
imaging radar provides metric range, elevation, return strength, and relative
radial motion, and remains effective in rain, fog, and night scenes. Recent
radar--camera detectors exploit camera--radar complementarity by aligning image features
with radar pillars, BEV grids, or object queries
\citep{RCFusion,MSSF,RCBEVDet,zfusion,HGSFusion,racformer}. Although the aligned
features make complementary cues available, scene-level alignment determines
\emph{where} evidence is placed in a shared representation but not
\emph{whether} it is admissible for the same evolving 3D hypothesis. We argue
that this second decision---evidence--hypothesis compatibility---should be
explicit at every object-level read.

Current fusion designs leave evidence--hypothesis consistency implicit. Dense BEV fusion
projects both modalities into a shared ground-plane representation
\citep{RCBEVDet,zfusion}, while query- and proposal-level methods aggregate
instance features after cross-modal association
\citep{CRAFT,racformer,CVFusion2025}. Neither design verifies evidence against
the evolving query's 3D radar geometry. A nearby image token may describe an
occluder or background, while a radar return may arise from multipath or a
reflector outside the object. Projection and proximity alone therefore cannot
determine which evidence supports the query.

Unlike LiDAR, 4D radar provides sparse and noisy returns rather than dense
object surfaces. Treating radar returns as a dense semantic map can introduce
false support, whereas ignoring their object structure discards useful geometry
and motion. We instead treat each return as potential support for a specific
anchor rather than as unrestricted scene context. This view separates
candidate access from evidence admission: proximity proposes evidence, while
compatibility with the current hypothesis determines how it may affect a
decision.

\sgdet{} provides a useful starting point by combining
radar-corrected depth, semantic radar pillars, and localization-aware object
attention \citep{SGDet3D}. Yet its geometry and semantics are still
coupled mainly through scene-level BEV features and projected supports.
\sgdetpp{} extends geometry-guided fusion through
\emph{hypothesis-conditioned evidence grounding}. Each sparse query uses its
evolving 3D anchor to condition or verify radar evidence before semantic
retrieval, geometric refinement, and temporal inheritance. The same principle
thus governs three decoder decisions rather than introducing three unrelated
feature blocks.

\begin{figure*}[t]
    \centering
    \includegraphics[width=\textwidth]{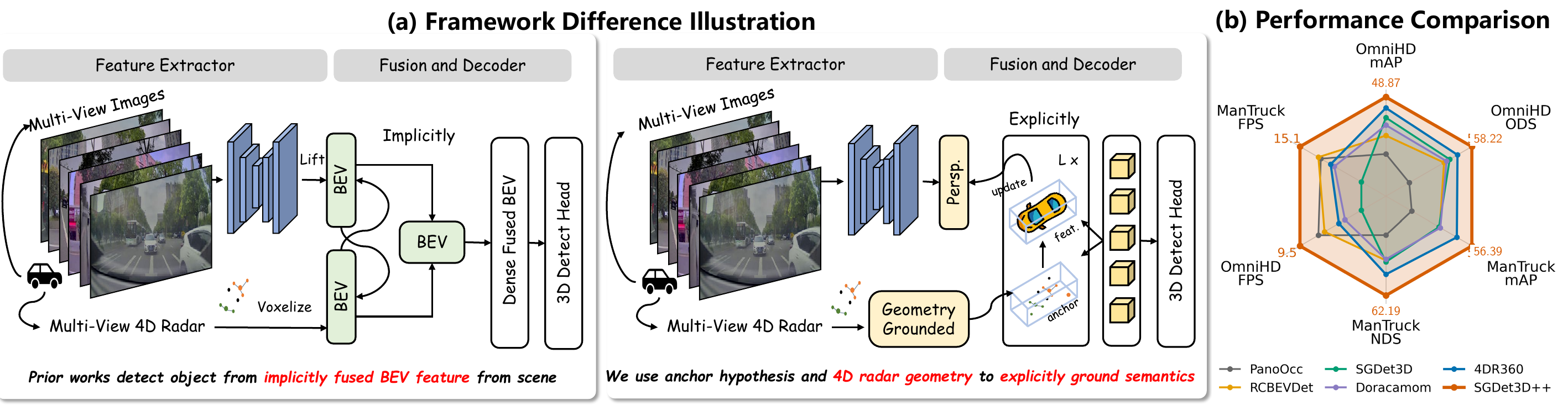}
    \caption{Conceptual and empirical comparison of radar--camera fusion
        strategies. \textit{(a)} Comparison between implicit BEV fusion and
        geometry-grounded semantics via hypothesis-conditioned evidence
        grounding in
        \sgdetpp{}.
        \textit{(b)} The radar chart compares representative methods on
        OmniHD-Scenes and ManTruckScenes and places \sgdetpp{} in their
        accuracy--efficiency landscape.}
    \label{fig:comparison}
    \vspace{-15pt}
\end{figure*}

Figure~\ref{fig:comparison}(a) compares scene-level implicit BEV fusion with
our hypothesis-conditioned design, in which each 3D anchor explicitly grounds
the evidence retrieved by its query. Figure~\ref{fig:comparison}(b)
compares the accuracy and efficiency of representative detectors on
OmniHD-Scenes and ManTruckScenes. Figure~\ref{fig:overview} summarizes the
\sgdetpp{} architecture. AGR, GCR, and DVC ground semantic retrieval, box
refinement, and temporal inheritance, respectively. At each decision, the
current anchor determines which radar evidence may influence the query.
Our contributions are summarized below.
\begin{itemize}
    \item We formulate \emph{hypothesis-conditioned evidence grounding} as a
        common object-decoder contract that separates candidate access from
        evidence admission: the evolving 3D state establishes semantic,
        geometric, or motion compatibility before evidence changes its query.
    \item We instantiate the principle at three decision interfaces. AGR uses
        pooled anchor-consistent radar statistics to condition image retrieval,
        GCR reads individual object-owned returns for box refinement, and DVC
        admits pose-aligned history with current radial motion.
    \item We validate accuracy and the proposed operating mechanism on three
        benchmarks. Controlled ablations isolate every module, while held-out
        decision-level diagnostics directly measure retrieval redistribution,
        return ownership, and motion-consistent memory inheritance.
\end{itemize}

\section{Related Work}
\label{sec:related}

\subsection{4D Radar-Based 3D Detection}
\label{sec:rw_radar}

Radar-only detectors encode sparse 4D-radar returns as points, voxels, or
pillars. Point-feature aggregation and multi-view fusion improve local
geometric descriptors \citep{RPFANet,MVFAN,MAFF-Net,DADAN}, while
pillar-oriented studies examine how sparse return support should be organized
for detection \citep{pillars-for-radar}. Local--global interaction and stronger
encoders further recover context from sparse measurements
\citep{LGDD,HyperDet2026,SD4R2025}. Radar nevertheless provides limited class
semantics and incomplete object boundaries. We therefore use its range,
reflectivity, and motion as object-level evidence rather than as a dense
semantic substitute for images.

\subsection{4D Radar and Camera Fusion 3D Detection}
\label{sec:rw_fusion}

Radar--camera detectors commonly align the modalities in perspective space or
a shared BEV representation. Projection and common BEV features complement
image appearance with radar range
\citep{RADIANT,RCFusion,CRN,LXL,RCMFusion,cr3dt,disadet,wavelet}. Later methods strengthen
BEV-level interaction with radar-aware fusion and multi-scale cross attention
\citep{RCBEVDet,zfusion,UniBEVFusion,HyDRa,RCGeoCP2026}. Other studies address
radar sparsity through sampling, generation, and synchronization
\citep{MSSF,HGSFusion}.

CRAFT associates radar returns with image proposals \citep{CRAFT}, RaCFormer
samples instance-relevant image and BEV features with object queries
\citep{racformer}, and CVFusion aggregates point, image, and BEV features for
proposal refinement \citep{CVFusion2025}. In contrast, \sgdetpp{} explicitly
associates each query with anchor-consistent radar geometry before semantic
retrieval, return-level refinement, and Doppler-verified temporal inheritance,
extending \sgdet{} \citep{SGDet3D} from scene-level fusion to explicit
object-level decisions.

\vspace{-10pt}
\section{Method}
\label{sec:method}

\subsection{Overview}
\label{sec:method_overview}

For each frame indexed by $t$, the detector receives synchronized multi-view
images, 4D radar returns, camera calibration, and ego poses. The detector
predicts 3D boxes with class labels and planar velocities, while depth serves
only as an auxiliary geometric signal for feature construction and training.
We write the input as
\begin{equation}
    \mathcal{X}_t=\{I_t^k\}_{k=1}^{N_c}\cup\mathcal{R}_t\cup\mathcal{C}_t
    \cup\mathcal{T}_t,
    \qquad
    \mathcal{Y}_t=\{(b_n^{\mathrm{gt}},c_n)\}_{n=1}^{N_t^{\mathrm{gt}}},
    \label{eq:problem_input}
\end{equation}
where $N_c$ is the number of cameras and $I_t^k$ is the image from camera $k$.
The sets $\mathcal{R}_t$, $\mathcal{C}_t$, and $\mathcal{T}_t$ contain the 4D
radar returns, camera calibration parameters, and ego poses and timestamps for
frame $t$, respectively. The target set contains $N_t^{\mathrm{gt}}$ objects.
For object $n$, $b_n^{\mathrm{gt}}$ specifies the 3D location, extent, yaw, and planar velocity
of object $n$, and $c_n$ is its class label. \sgdetpp{} realizes
geometry-grounded semantics by associating radar support with each evolving
hypothesis before semantic retrieval, geometric refinement, or temporal
inheritance.
% The decoder may additionally
% predict an auxiliary quality vector $u_i=(u_i^{\mathrm{ctr}},u_i^{\mathrm{yaw}})$
% for centre and yaw quality; this vector is used by the refinement losses and
% is not a separate element of the detection target.

\begin{figure*}[t]
    \centering
    \begin{tikzpicture}
        \node[anchor=south west,inner sep=0] (framework) at (0,0)
        {\includegraphics[width=\textwidth]{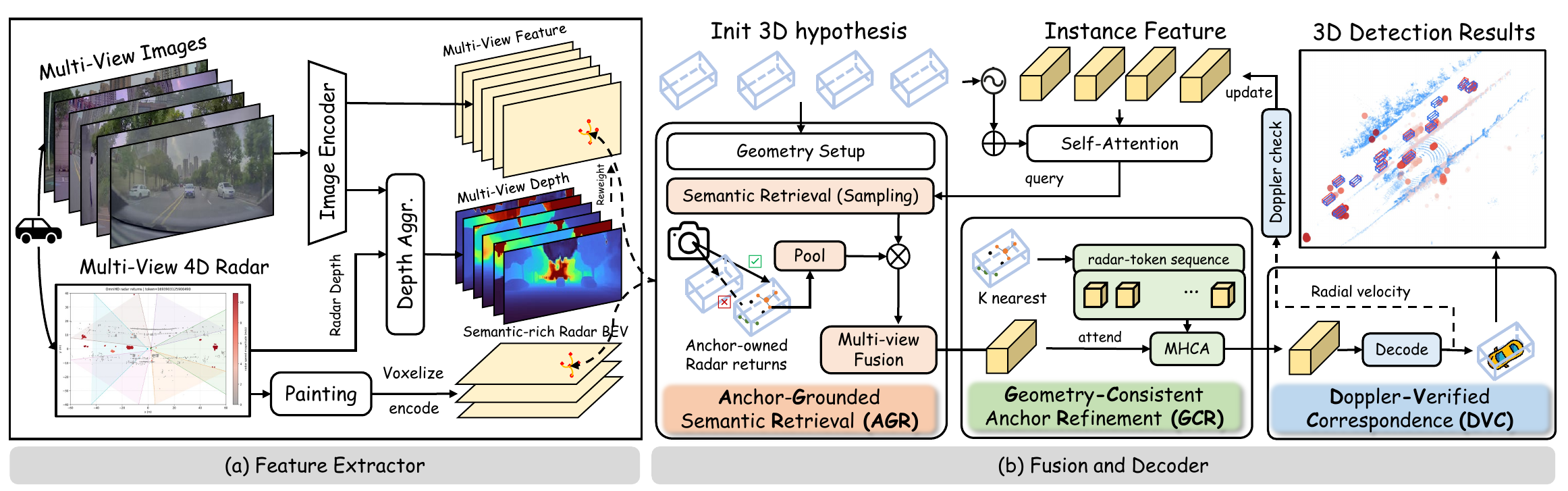}};
        \begin{scope}[x={(framework.south east)},y={(framework.north west)}]
            \node[fill=white,align=center,inner sep=0.6pt]
            at (0.470,0.285)
            {\resizebox{0.073\textwidth}{!}{\footnotesize
                    \shortstack{Anchor-consistent\\radar support}}};
        \end{scope}
    \end{tikzpicture}
    \caption{\sgdetpp{} grounds object-query decoding in sparse 4D-radar
        evidence. AGR uses anchor-consistent radar support to reweight image tokens at
        box-conditioned supports, GCR supplies query-weighted local return evidence
        for box refinement, and DVC uses current radial Doppler to verify whether a
        historical object slot should be retained or replaced.}
    \label{fig:overview}
    \vspace{-20pt}
\end{figure*}

As shown in Figure~\ref{fig:overview}, \sgdetpp{} builds on the
semantic--geometric principle of \sgdet{} \citep{SGDet3D}, replacing dense
image-to-BEV refinement with sparse object queries that sample image features
at box-conditioned supports. Corrected-depth painting \citep{SARCD} enriches occupied radar
pillars, which the radar encoder scatters into the BEV feature map $F_t^r$.
The image encoder produces the multi-camera, multi-level feature pyramid
$F_t^I$. The front end couples image semantics
with radar geometry, and the sparse object decoder grounds its evidence in the
geometry of each object hypothesis. At layer
$\ell$, each anchor reads $F_t^I$ and $F_t^r$ through box-conditioned supports
and updates its state according to
\begin{equation}
    (f_i^{\ell+1},b_i^{\ell+1})
    =\mathcal{D}^{\ell}\bigl(
    f_i^\ell,b_i^\ell,F_t^I,F_t^r,\mathcal{R}_t,\mathcal{M}_{t-1}\bigr),
    \label{eq:decoder_overview}
\end{equation}
where $f_i^\ell$ and $b_i^\ell$ are the feature and 3D box state of query $i$
before layer $\ell$, and $\mathcal{M}_{t-1}$ is the object memory carried from
the preceding frame. The operator $\mathcal{D}^{\ell}$ denotes the complete
update performed by decoder layer $\ell$ from the query state, image and radar
features, raw radar returns, and temporal memory shown in
Eq.~\ref{eq:decoder_overview}. AGR, GCR, and DVC perform semantic, geometric,
and temporal grounding within the decoder update. For every operation, the
current anchor defines radar support before the selected evidence modifies the
object state. The decoder therefore selects evidence for each object instead
of introducing another fused scene representation.
% AGR
% conditions image retrieval on anchor-consistent radar evidence. GCR aggregates
% multiple local returns through object-to-return attention before box
% refinement. DVC verifies pose-aligned historical anchors with current radial
% Doppler before their features are inherited.
\vspace{-5pt}
\subsection{Hypothesis-Conditioned Evidence Grounding}
\label{sec:object_refinement}

Let $h_i^\ell=(f_i^\ell,b_i^\ell)$ denote the feature and 3D state of
hypothesis $i$. We define evidence grounding as an operation-specific map
\begin{equation}
    e_i^{m,\ell}=\Gamma_m\!\left(E_t^m;h_i^\ell\right),
    \qquad m\in\{\mathrm{sem},\mathrm{geo},\mathrm{temp}\},
    \label{eq:hypothesis_grounding}
\end{equation}
where $E_t^m$ is the candidate evidence available to a semantic, geometric, or
temporal decoder operation. Unlike unrestricted feature fusion,
$\Gamma_m$ must expose a physical compatibility relation between the candidate
and the consuming hypothesis. AGR realizes $\Gamma_{\mathrm{sem}}$ by
conditioning image-retrieval weights on anchor-consistent return statistics;
GCR realizes $\Gamma_{\mathrm{geo}}$ by retaining box-owned returns as
individual tokens; and DVC realizes $\Gamma_{\mathrm{temp}}$ by testing a
history slot against current radial motion. This shared decision interface
supports evidence-specific operators under a common evidence--hypothesis
compatibility contract.

The decoder maintains $N_q$ object hypotheses. At layer $\ell$, hypothesis $i$
has feature $f_i^\ell$ and box state
$b_i^\ell=(x_i^\ell,y_i^\ell,z_i^\ell,w_i^\ell,l_i^\ell,h_i^\ell,
\theta_i^\ell,v_{x,i}^\ell,v_{y,i}^\ell)$. The first three components define
the box centre $o_i^\ell=(x_i^\ell,y_i^\ell,z_i^\ell)$. The next three give
its width, length, and height, while the remaining components specify yaw and
planar velocity.
The current box generates 13 supports, including its centre, six fixed surface
supports, and six learned object-relative supports. Let $s_j\in\mathbb{R}^3$
be the normalized object-relative offset of support $j$, let
$R(\theta_i^\ell)$ be the rotation about the vertical axis defined by the
current yaw, and let $\odot$ denote element-wise multiplication. Each support
location is
\begin{equation}
    p_{ij}^\ell=o_i^\ell+
    R(\theta_i^\ell)\bigl(s_j\odot(w_i^\ell,l_i^\ell,h_i^\ell)\bigr),
    \qquad j=1,\ldots,13.
    \label{eq:box_supports}
\end{equation}
\begin{wrapfigure}{l}{0.56\columnwidth}
    \centering
    \vspace{-10pt}
    \includegraphics[width=\linewidth]{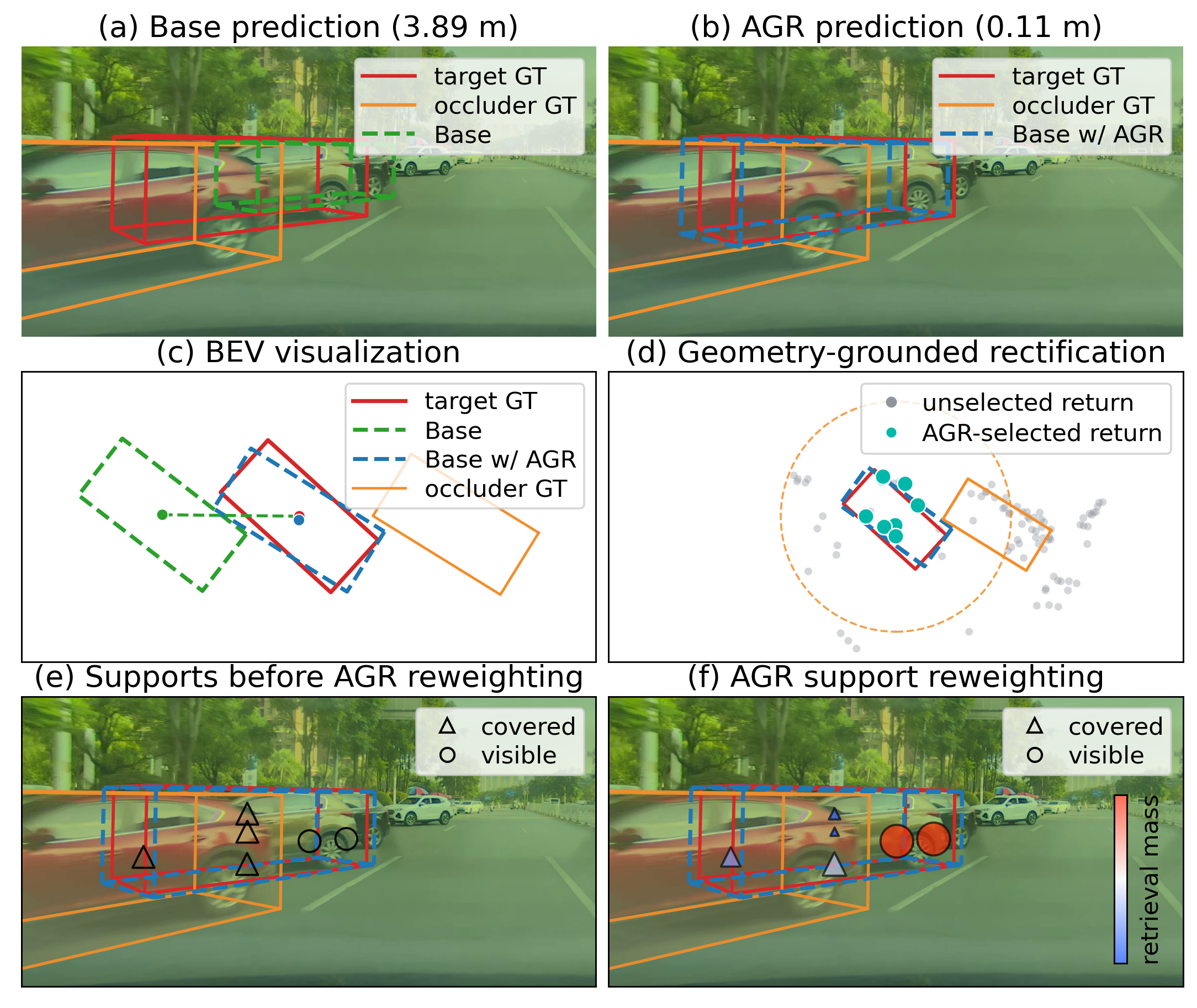}
    \vspace{-15pt}
    \caption{Qualitative AGR evidence in an occluded scene. (a--c) AGR reduces
        the target center error relative to Base. (d) The yaw-aligned box and
        radius gates restrict returns by anchor extent and association range;
        their intersection defines the anchor-consistent radar support.
        (e--f) Gated support shifts retrieval mass from occluder-covered
        supports (triangles) to visible supports (circles). Color indicates
        signed change from decrease (blue) to increase (red), and marker area
        its magnitude.}
    \label{fig:ground_issue}
    \vspace{-25pt}
\end{wrapfigure}
Each location $p_{ij}^\ell$ samples multi-view FPN features and the radar BEV feature map
through deformable feature fusion. The subsequent self-attention, feed-forward,
and refinement operations update both $f_i^\ell$ and $b_i^\ell$. Thus, the
location and extent of an anchor determine the accessible evidence rather
than retrieving the same undifferentiated scene representation for every query.

The decoder's self-attention blocks use decoupled multi-head attention. Query
and key projections include anchor embeddings, whereas value projections carry
object features. Although decoupled attention allows a current query to
interact with other current queries and pose-aligned historical queries, the
attention mechanism does not establish whether a particular radar return
supports that query. AGR and GCR
therefore apply the association rule in Eq.~\ref{eq:anchor_consistent_returns}
before image-feature retrieval and box refinement.

\noindent\textbf{Anchor-Grounded Semantic Retrieval (AGR).}
AGR uses anchor-consistent radar support to guide image retrieval.
Figure~\ref{fig:ground_issue} illustrates the effect of its two association gates.
The yaw-aligned box gate removes returns outside the current anchor, and the
radius gate prevents distant returns from entering the pooled radar descriptor
(Figure~\ref{fig:ground_issue}(d)). Using the returns
that pass both gates shifts retrieval mass from occluder-covered to visible
supports in this case (Figure~\ref{fig:ground_issue}(e--f)) and yields the
better-localized box in Figure~\ref{fig:ground_issue}(a--c). The aggregate
evaluation quantifies retrieval redistribution and occlusion-stratified
detection across the validation set.
For an anchor $b_i^\ell$, we write $r$ for the 3D position of a radar return and
transform the return into the yaw-aligned local coordinate
$\widetilde r_{ir}=R(\theta_i^\ell)^\top(r-o_i^\ell)$. Let
$u_i^\ell=(w_i^\ell,l_i^\ell,h_i^\ell)$ be the anchor extent. With anchor
centre $o_i^\ell$, margin $\delta$, and radius $\rho$, AGR assigns return
$r$ to anchor $i$ only if
\begin{equation}
    \mathcal{R}_i^\ell=\left\{r\,\middle|\,
    \left|R(\theta_i^\ell)^\top(r-o_i^\ell)\right|
    \preceq \tfrac{1}{2}u_i^\ell+\delta,
    \quad \|r-o_i^\ell\|_2\leq\rho\right\}.
    \label{eq:anchor_consistent_returns}
\end{equation}
The absolute value and inequality $\preceq$ are applied component-wise, so the
first condition is a two-sided membership test in the yaw-aligned 3D box. The
extent gate excludes returns outside the current anchor, while the radius gate
prevents large or uncertain anchors from collecting distant returns. 
We call the selected set \emph{anchor-consistent radar support}. 

AGR pools the associated returns into an anchor-level evidence vector. For
$n_i=|\mathcal{R}_i^\ell|>0$, let $P$ be the input return budget and use an
overbar for the mean over $\mathcal{R}_i^\ell$. Let $S(r)$,
$\operatorname{qual}(r)$, and $\operatorname{age}(r)$ denote the return
strength, quality score, and sweep age of return $r$, respectively. The
seven-dimensional descriptor, its $D$-dimensional embedding, and the resulting
query update are
\begin{equation}
    \begin{aligned}
        d_i^\ell
                 & =\left[
            \overline{\widetilde r}_{i,x},
            \overline{\widetilde r}_{i,y},
            \overline{\widetilde r}_{i,z},
            \frac{\log n_i}{\log P},
            \overline{\frac{S(r)}{\max(\|r\|_2,1\,\mathrm{m})^2}},
            \overline{\operatorname{qual}(r)},
            \overline{\operatorname{age}(r)}
        \right]\in\mathbb{R}^{7}, \\
        e_i^\ell
            &=\Phi_{\mathrm{AGR}}(d_i^\ell)\in\mathbb{R}^{D}, \qquad
        \widehat f_i^\ell
            =f_i^\ell+\Phi_q([f_i^\ell\Vert e_i^\ell]).
    \end{aligned}
    \label{eq:agr_evidence_update}
\end{equation}
Here $P=768$ is the maximum number of input returns and $D=256$ is the decoder
feature dimension in all reported experiments. The symbol $\Vert$ denotes
feature concatenation, while $\Phi_{\mathrm{AGR}}$ and $\Phi_q$ are learned
MLPs that encode the pooled radar descriptor and compute its residual query
update, respectively. The three local-coordinate
means remain in metres. The count is logarithmically normalized to $[0,1]$,
and return strength $S(r)$---radar power on OmniHD-Scenes and RCS on
ManTruckScenes---is normalized by the squared sensor range, clipped below at
1\,m. The quality and sweep-age channels retain the shared loader's native
units and are zero-filled when unavailable. The stated normalization keeps each
physical cue on a documented scale before embedding.

The updated query $\widehat f_i^\ell$ reweights the projected image tokens without changing
the 13 support locations $p_{ij}^\ell$ from Eq.~\ref{eq:box_supports}.
\begin{equation}
    \omega_{icsjg}
    = \operatorname{softmax}_{(c,s,j)}
    \left(
    \left[\Phi_{\omega}
        \left(\widehat f_i^\ell+\alpha_i^\ell+\gamma_c\right)\right]_{csjg}
    \right).
    \label{eq:agr_image_weights}
\end{equation}
Here $c$, $s$, $j$, and $g$ index the camera, FPN level, anchor support, and
feature group, respectively. The vectors $\alpha_i^\ell$ and $\gamma_c$ are
the learned embeddings of anchor $i$ and camera $c$, and $\Phi_\omega$ maps the
combined query to one sampling logit for each $(c,s,j,g)$ tuple. For every
feature group $g$, the softmax normalizes the logits over all camera--level--support
triples $(c,s,j)$. AGR then aggregates the corresponding image features as
\begin{equation}
    f_{i}^{I,\ell}
    =\sum_g\sum_{c,s,j}\omega_{icsjg}
    W_gF_{s,c}^{I}\!\left(\pi_c(p_{ij}^\ell)\right).
    \label{eq:agr_image_retrieval}
\end{equation}
Here $F^I_{s,c}$ is the image feature map from camera $c$ at FPN level $s$,
$\pi_c$ is the calibrated projection from a 3D point to camera $c$, and $W_g$
projects the sampled feature into group $g$. AGR changes the contribution of
each image token after anchor-consistent association while leaving the support
locations fixed and introducing no dense radar branch or direct box correction.
GCR then uses the same associated return set for box refinement while preserving
selected returns as individual evidence.
% \begin{figure*}[t]
%     \centering
%     \includegraphics[width=0.80\textwidth]{Fig4-grounded-semantics-placeholder.pdf}
%     \caption{Placeholder for sample-level geometry-grounded semantic
%     visualization.}
%     \label{fig:grounded_semantics}
%     \vspace{-20pt}
%     % Unlike the framework diagram in Figure~\ref{fig:overview},
%     % this figure
%     % will use real scenes to show how image semantics are accepted only when
%     % they correspond to anchor-consistent 4D-radar geometry. The intended panels show
%     % image-space semantic evidence and anchor-consistent BEV radar support, and
%     % the resulting geometry-grounded semantic correspondence.
% \end{figure*}

\noindent\textbf{Geometry-Consistent Anchor Refinement (GCR).}
After semantic retrieval, box refinement still faces sparse and noisy radar
evidence. A valid object may produce only a few returns from different
surfaces, with varying return strength and Doppler reliability.
Even after the association gate in
Eq.~\ref{eq:anchor_consistent_returns}, not every retained return is equally
useful for correcting the box. GCR therefore keeps several associated returns
and lets the current object feature attentively select the radar evidence used
for anchor refinement.

Starting from $\mathcal{R}_i^\ell$, GCR keeps up to $K_r$ nearest associated returns
as $\mathcal{K}_i^\ell$. Each return is encoded with the cues needed for
refinement.
\begin{equation}
    \begin{aligned}
        h_{ir}=\Phi_{\mathrm{ret}}\!\Big(
        \big[
            \underbrace{
            \widetilde r_{ir}/\rho
            \Vert
            \|\widetilde r_{ir}\|_2/\rho
            }_{\text{anchor-local geometry}}
            \Vert
            \underbrace{
                S(r)/d(r)^2
                \Vert
                \operatorname{qual}(r)
                \Vert
                \operatorname{age}(r)
                \Vert
                \nu(r)
            }_{\text{signal and radial motion}}
            \big]\Big),
        \quad r\in\mathcal{K}_i^\ell .
    \end{aligned}
    \label{eq:gcr_return_evidence}
\end{equation}
Here $\Phi_{\mathrm{ret}}$ is the learned return encoder, $\rho$ is the support
radius, $d(r)=\|r\|_2$ is the sensor-to-return range, and $\nu(r)$ is the radial
Doppler measurement. The quality and age functions have the same definitions
as in Eq.~\ref{eq:agr_evidence_update}. The anchor-local geometry
indicates where the return lies with respect to the current box, while return
strength, quality, sweep age, and Doppler describe the strength, reliability,
recency, and radial motion of the measurement. Keeping the cues at the return level lets the
model distinguish a strong surface return from a weak or motion-inconsistent
return that passes the coarse association gate without agreeing with the
current object state.

The encoded returns form a radar-token sequence, and the anchor feature
provides the query.
\begin{equation}
    \rmT_i^\ell=[h_{ir_1}^{\top},\ldots,h_{ir_{K_r}}^{\top}]^{\top},\qquad
    q_i^\ell=W_qf_i^\ell,\quad
    \rmK_i^\ell=W_k\rmT_i^\ell,\quad
    \rmV_i^\ell=W_v\rmT_i^\ell .
    \label{eq:gcr_radar_tokens}
\end{equation}
Here $\rmT_i^\ell$ is the return-token matrix, and $W_q$, $W_k$, and $W_v$ are
learned query, key, and value projections. The current object feature then
aggregates the sparse radar-token sequence by attention.
\begin{equation}
    \begin{aligned}
        \beta_i
         & = \operatorname{softmax}
        \left(
        \frac{q_i^\ell(\rmK_i^\ell)^\top}{\sqrt{D}}+m_i
        \right), \quad
        g_i^\ell
         & = \beta_i\rmV_i^\ell .
    \end{aligned}
    \label{eq:gcr_object_return_attention}
\end{equation}
Here $D$ is the feature dimension defined above. Element $m_{ir}$ of mask
$m_i$ is $0$ for a valid selected return and a large negative value for a
padded slot.
The attention weights $\beta_i$ let each anchor emphasize returns whose
local geometry, signal strength, quality, and radial motion agree with its
current object state, while suppressing less compatible returns. The resulting
radar context $g_i^\ell$ is concatenated with $f_i^\ell$ and injected before
the ordinary box refiner, so radar influences centre, size, orientation, and
velocity through the refinement representation. Return strength and Doppler are
therefore used as learnable evidence fields, not as hard thresholds or direct
box residuals. If no anchor-consistent radar support is available, the anchor
follows the original refinement path. AGR uses pooled radar evidence for
semantic retrieval, while GCR attends to individual returns for box refinement.

\begin{table*}[t]
    \belowrulesep=0pt
    \aboverulesep=0pt
    \caption{OmniHD-Scenes 3D detection results from our adapted benchmark
        suite.}
    \label{tab_omnihd_det}
    \centering
    \small
    \setlength{\tabcolsep}{1.5pt}
    \renewcommand\arraystretch{1.00}
    \resizebox{\textwidth}{!}{%
        \begin{tabular}{@{}L{3.10cm}|C{0.65cm}|C{1.00cm}C{1.00cm}|C{1.00cm}C{1.00cm}C{1.00cm}C{1.00cm}|C{1.00cm}C{1.00cm}C{1.00cm}C{1.00cm}|C{0.80cm}@{}}
            \toprule
            \makebox[3.10cm][l]{Method} & \makebox[0.65cm]{Mod.} &
            \makebox[1.00cm]{mAP} & \makebox[1.00cm]{ODS} &
            \makebox[1.00cm]{mATE} & \makebox[1.00cm]{mASE} &
            \makebox[1.00cm]{mAOE} & \makebox[1.00cm]{mAVE} &
            \makebox[1.00cm]{Car} & \makebox[1.00cm]{Ped.} &
            \makebox[1.00cm]{Rider} & \makebox[1.00cm]{LVeh.} &
            \makebox[0.80cm]{FPS} \\
            \midrule
            PointPillars'19 & R & 23.82 & 37.21 &
            0.6752 & 0.2447 & 0.3776 & 0.6789 &
            52.74 & 0.69 & 28.57 & 13.29 & 62.2 \\
            RadarPillarNet'23 & R & 24.88 & 37.81 &
            0.6597 & 0.2389 & 0.3736 & 0.6982 &
            52.99 & 2.06 & 29.45 & 15.02 & 60.3 \\
            \midrule
            BEVFormer'22 & C & 29.17 & 30.54 &
            1.1046 & 0.2346 & 0.4889 & 1.0797 &
            53.64 & 14.48 & 33.55 & 15.01 & 11.4 \\
            PanoOcc'24 & C & 29.17 & 28.55 &
            1.1500 & 0.2446 & 0.6378 & 1.6066 &
            51.58 & 15.82 & 35.02 & 14.26 & 5.5 \\
            \midrule
            BEVFusion'22 & R+C & 33.95 & 42.62 &
            0.5730 & 0.2465 & 0.3814 & 0.7474 &
            56.25 & 11.66 & 50.90 & 16.99 & 3.6 \\
            RCFusion'23 & R+C & 34.88 & 40.65 &
            0.5676 & 0.2535 & 0.4011 & 0.9208 &
            57.17 & 12.87 & 51.35 & 18.11 & 3.6 \\
            RCBEVDet'24 & R+C & 35.53 & 45.04 &
            \thirderrcell{0.5138} & \seconderrcell{0.2305} &
            0.3914 & 0.6825 &
            62.35 & 10.11 & 54.60 & 15.06 & \secondcell{5.2} \\
            SGDet3D'25 & R+C & \thirdcell{41.73} & \thirdcell{47.70} &
            0.5430 & 0.2369 & 0.3885 & 0.6849 &
            \thirdcell{63.30} & 19.00 & \thirdcell{58.30} &
            \thirdcell{26.30} & 3.4 \\
            RaGS'26 & R+C & 35.88 & 43.45 &
            \na & \na & \na & \na &
            \na & \na & \na & \na & \na \\
            Doracamom-S'26 & R+C & 37.60 & 41.31 &
            0.6724 & \thirderrcell{0.2329} & 0.4359 & 0.8579 &
            58.94 & 17.84 & 52.72 & 20.89 & \thirdcell{4.8} \\
            Doracamom'26 & R+C & 39.12 & 46.22 &
            0.6646 & 0.2331 & \seconderrcell{0.3545} &
            \thirderrcell{0.6151} &
            61.12 & 19.83 & 53.35 & 22.18 & 4.2 \\
            4DR360$^\circ$'26 & R+C &
            \secondcell{45.05} & \secondcell{51.40} &
            \seconderrcell{0.4705} & 0.2423 & 0.3762 &
            \seconderrcell{0.6013} &
            \secondcell{65.47} & \secondcell{21.73} &
            \secondcell{60.41} & \secondcell{32.58} & 4.5 \\
            \cellcolor{gray!20} \sgdetpp{} \textbf{(ours)} &
            \cellcolor{gray!20} R+C & \bestcell{48.87} & \bestcell{58.22} &
            \besterrcell{0.4298} & \besterrcell{0.1968} &
            \besterrcell{0.3355} & \besterrcell{0.3350} &
            \bestcell{72.02} & \bestcell{24.50} & \bestcell{62.17} &
            \bestcell{36.78} & \bestcell{9.5} \\
            \bottomrule
        \end{tabular}%
    }
    \vspace{-10pt}
\end{table*}

\begin{table*}[t]
    \belowrulesep=0pt
    \aboverulesep=0pt
    \caption{ManTruckScenes 3D detection results.}
    \label{tab_mantruck_det}
    \centering
    \small
    \setlength{\tabcolsep}{1.5pt}
    \renewcommand\arraystretch{1.00}
    \resizebox{\textwidth}{!}{%
        \begin{tabular}{@{}L{3.10cm}|C{0.65cm}|C{1.00cm}C{1.00cm}|C{1.00cm}C{1.00cm}C{1.00cm}C{1.00cm}|C{1.00cm}C{1.00cm}C{1.00cm}C{1.00cm}|C{0.80cm}@{}}
            \toprule
            \makebox[3.10cm][l]{Method} & \makebox[0.65cm]{Mod.} &
            \makebox[1.00cm]{mAP} & \makebox[1.00cm]{NDS} &
            \makebox[1.00cm]{mATE} & \makebox[1.00cm]{mASE} &
            \makebox[1.00cm]{mAOE} & \makebox[1.00cm]{mAVE} &
            \makebox[1.00cm]{Car} & \makebox[1.00cm]{LVeh.} &
            \makebox[1.00cm]{Trailer} & \makebox[1.00cm]{Obs.} &
            \makebox[0.80cm]{FPS} \\
            \midrule
            RadarPillarNet'23 & R & 26.61 & 37.73 &
            0.5243 & 0.2413 & 0.1689 & 2.1915 &
            44.43 & 30.66 & 28.65 & 2.68 & 64.2 \\
            LGDD'25 & R & 29.83 & 40.49 &
            0.4922 & 0.2328 & 0.1229 & 1.5775 &
            48.90 & 36.20 & 32.40 & 1.80 & 20.1 \\
            \midrule
            BEVFormer'22 & C & 28.30 & 35.47 &
            0.7662 & 0.2583 & 0.1982 & 1.2747 &
            39.56 & 28.10 & 19.66 & 25.87 & 10.7 \\
            PanoOcc'24 & C & 28.82 & 35.72 &
            0.7686 & 0.2653 & 0.1920 & 1.2623 &
            40.12 & 28.22 & 21.73 & 25.20 & 6.3 \\
            \midrule
            BEVFusion'22 & R+C & 30.97 & 40.65 &
            0.5168 & 0.2486 & 0.1249 & 2.1377 &
            47.24 & 36.09 & 30.15 & 10.40 & 4.6 \\
            RCFusion'23 & R+C & 31.61 & 40.57 &
            0.5652 & 0.2483 & \seconderrcell{0.1155} & 2.0670 &
            48.51 & 36.84 & 28.58 & 12.49 & 5.4 \\
            LXL'24 & R+C & 34.70 & 43.04 &
            0.4972 & 0.2470 & \thirderrcell{0.1174} & 1.9877 &
            51.95 & 41.14 & 31.41 & 14.30 & 5.2 \\
            RCBEVDet'24 & R+C & 41.72 & 46.85 &
            0.4500 & 0.2658 & 0.1543 & \thirderrcell{1.0407} &
            \secondcell{61.93} & 38.00 & 34.38 & 32.58 &
            \secondcell{6.5} \\
            HGSFusion'25 & R+C & \thirdcell{43.90} & \thirdcell{48.23} &
            0.4607 & \thirderrcell{0.2272} & 0.1668 & 1.9479 &
            58.76 & 43.64 & \thirdcell{35.43} & \thirdcell{37.77} & 5.1 \\
            SGDet3D'25$^\dagger$ & R+C & 41.37 & 47.50 &
            \thirderrcell{0.4252} & \seconderrcell{0.2159} &
            0.1523 & 1.6112 &
            60.82 & \secondcell{47.38} & 26.94 & 30.29 & 3.0 \\
            Doracamom'26 & R+C & 41.98 & 46.36 &
            0.4874 & 0.2411 & 0.1982 & 1.5566 &
            \thirdcell{61.62} & 42.71 & 34.45 & 29.15 & 5.2 \\
            4DR360$^\circ$'26 & R+C &
            \secondcell{49.57} & \secondcell{52.97} &
            \besterrcell{0.3613} & 0.2485 & 0.1571 &
            \seconderrcell{0.9442} &
            59.41 & \thirdcell{44.71} & \secondcell{41.20} &
            \secondcell{52.97} & \thirdcell{5.5} \\
            \cellcolor{gray!20} \sgdetpp{} \textbf{(ours)} &
            \cellcolor{gray!20} R+C & \bestcell{56.39} & \bestcell{62.19} &
            \seconderrcell{0.3637} & \besterrcell{0.2025} &
            \besterrcell{0.0814} & \besterrcell{0.5749} &
            \bestcell{70.85} & \bestcell{48.32} & \bestcell{43.18} &
            \bestcell{63.24} & \bestcell{15.1} \\
            \bottomrule
        \end{tabular}%
    }
    \vspace{-20pt}
\end{table*}

\noindent\textbf{Doppler-Verified Correspondence (DVC).}
DVC verifies temporal memory with current Doppler evidence. The decoder stores
high-confidence object features and anchors, then transforms historical anchors
to the current frame using ego pose and timestamp. Although ego-pose
transformation removes displacement caused by ego motion, a stale or incorrect
memory slot may remain spatially close to a current query. DVC therefore checks
whether nearby radar returns have \emph{relative radial} motion consistent with
the motion predicted by the aligned historical anchor.

All quantities below are expressed in the current ego frame. For a return at
3D position $r$, define its unit line-of-sight direction as
$\widehat d_r=r/\|r\|_2$. For a pose-aligned historical anchor $a$
with centre $c_a$ and ego-motion-compensated object velocity
$v_a=(v_{x,a},v_{y,a})$ resolved in the current ego axes, DVC first
gathers current returns within a local support radius.
\begin{equation}
    \mathcal{Q}_a
    =\left\{r\in\mathcal{R}_t\,\middle|\,
    \|r-c_a\|_2\leq\rho_{\mathrm{DVC}}\right\}.
    \label{eq:dvc_local_returns}
\end{equation}
When $\mathcal{Q}_a$ is non-empty, DVC selects the spatially closest return
$r_a^*=\arg\min_{r\in\mathcal{Q}_a}\|r-c_a\|_2$. DVC projects the anchor
velocity onto the return direction as
\begin{equation}
    \widehat\nu_a
    =
    \left\langle
    [v_{x,a},v_{y,a},0]^\top,
    \widehat d_{r_a^*}
    \right\rangle .
    \label{eq:dvc_radial_prediction}
\end{equation}
The inner product $\langle\cdot,\cdot\rangle$ in
Eq.~\ref{eq:dvc_radial_prediction} projects the anchor's planar velocity onto
the selected return's line of sight, yielding the predicted radial velocity
$\widehat\nu_a$. The value $\nu(r_a^*)$ is the loader-provided,
ego-motion-compensated radial velocity of the selected return. DVC therefore
compares two radial quantities rather than interpreting Doppler as an
unconstrained BEV velocity. DVC marks a historical anchor as contradicted only
when a nearby return exists and the radial-velocity mismatch exceeds threshold
$\tau_v$. The resulting test is
\begin{equation}
    z_a=\mathbb{1}\!\left[
        \mathcal{Q}_a\neq\varnothing\ \land\
        \left|\widehat\nu_a-\nu(r_a^*)\right|>\tau_v
        \right].
    \label{eq:dvc_doppler_support}
\end{equation}
where $\mathbb{1}[\cdot]$ is the indicator function and $z_a=1$ denotes a
contradiction. The memory contains $M$ historical slots, which share bank
indices with the first $M$ current-query slots. If history slot $k$ is
contradicted, DVC substitutes the $k$-th freshly
initialized current feature and anchor. Otherwise, DVC retains the history pair.
The deterministic mapping uses neither nearest-object nor Hungarian matching.
Missing radar evidence preserves history, and the update adds no learnable
parameters while editing the existing temporal sequence without a second
cache.

\subsection{Detection Objective}
\label{sec:loss}

The detection head supervises classification, box regression, centre quality,
yaw quality, and planar velocity at decoder refinement layers. With
$\mathcal{L}_{\mathrm{det}}$ denoting the weighted detection objective and
$\mathcal{L}_{\mathrm{depth}}$ the multi-scale corrected-depth objective, the
training loss is
\begin{equation}
    \mathcal{L}=\mathcal{L}_{\mathrm{det}}+
    \lambda_{\mathrm{depth}}\mathcal{L}_{\mathrm{depth}}.
    \label{eq:loss}
\end{equation}
% The depth term provides geometric supervision for semantic addressing; all
% reported outputs and metrics are 3D detection quantities.

\section{Experiments}
\label{sec:experiments}

% Beyond overall accuracy, matched controls test four predictions. Box-aware
% association improves over radius-only association, AGR helps under projected
% occlusion, return-level cues improve refinement, and DVC improves fixed-budget
% ordinary memory.

% We evaluate \sgdetpp{} as a detection-only 4D radar--camera detector. The
% experiments are organized to test the central claim of this paper, namely
% that image semantics should be accessed only through object-level 4D radar
% geometry.
% Accordingly, we first compare against radar-only, camera-only, and R+C
% detectors on two benchmarks, then test robustness under degraded visual or
% calibration conditions, and finally isolate AGR, GCR, and DVC.

\begin{figure*}[t]
    \centering
    \includegraphics[width=\textwidth]{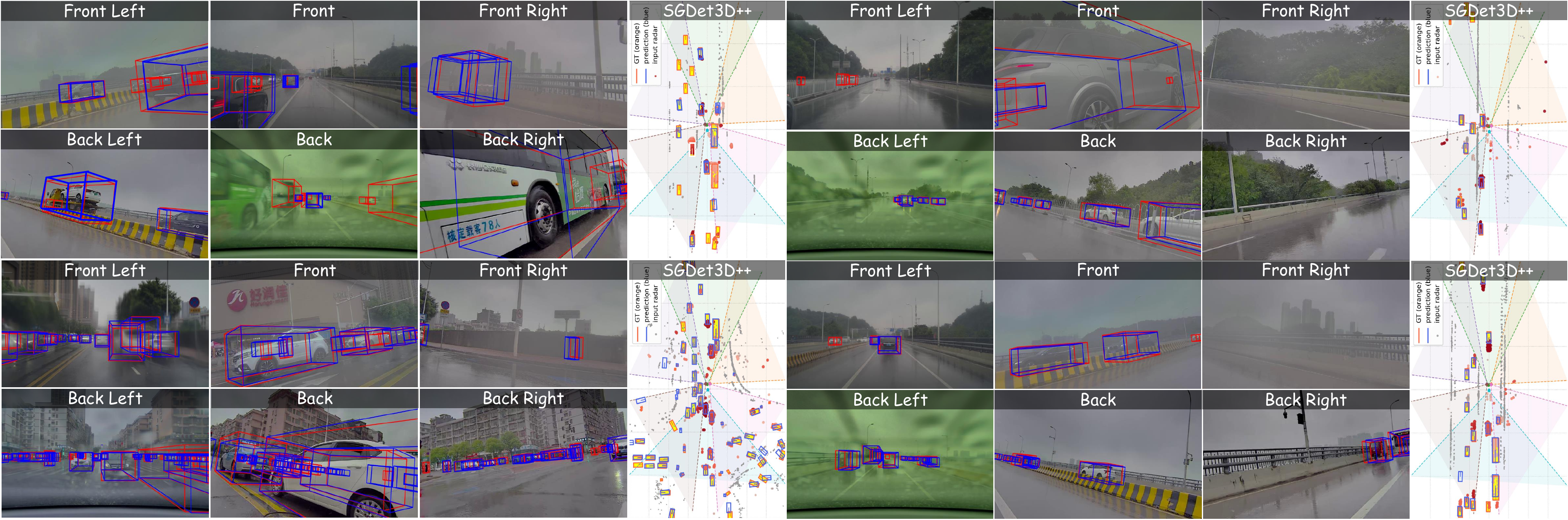}
    \caption{Qualitative visualization of \sgdetpp{} on
        OmniHD-Scenes. Each example pairs multi-view image detections with BEV
        radar evidence colored by radial-speed magnitude. Red boxes denote
        ground truth and blue boxes denote predictions.}
    \label{fig:qualitative_results}
    \vspace{-15pt}
\end{figure*}

\noindent\textbf{Datasets and metrics.}
We evaluate 3D detection on OmniHD-Scenes \citep{OmniHD} and ManTruckScenes
\citep{Man-TruckScenes}. OmniHD-Scenes pairs six cameras with 4D radar and
reports mAP, object detection score (ODS), four error metrics, and class AP for
car, pedestrian, rider, and large vehicle. ManTruckScenes complements it with
commercial-vehicle scenes and reports mAP, NDS, the same errors, and AP for
car, large vehicle, trailer, and obstacle.
% We also evaluate
% an adverse-condition OmniHD-Scenes subset and calibration perturbations to
% measure whether semantic retrieval remains supported by geometry when visual
% or projection evidence becomes unreliable.

\noindent\textbf{Implementation details.}
We train in MMDetection3D with AdamW and mixed precision on four RTX PRO 6000
GPUs. Within each dataset, all adapted methods share the validation split,
synchronized inputs, spatial range, and official evaluator. The ManTruckScenes \sgdet{} result marked $\dagger$ is from
4DR360$^\circ$ \citep{DR3602026}; the remaining \sgdet{} results use our adapted
control. ``Base'' disables AGR, GCR, and DVC. The \sgdet{} benchmark row reports
one designated run; the controlled Base result in
Table~\ref{tab:ablation_ALL} is instead the mean of three independent runs.
% It retains the ordinary recurrent object memory.

% \noindent\textbf{Comparison protocol and model identities.}
% Within each dataset, all adapted evaluations use the same split, synchronized
% sensor-input contract, spatial range, and official evaluator. We preserve method-native
% backbones, pretraining, fusion blocks, and decoders where applicable; the main
% tables are therefore source-faithful benchmark comparisons rather than
% architecture-matched ablations. In particular, the calibration control is
% the adapted SGDet3D evaluated under extrinsic perturbations. ``Base'' in
% Table~\ref{tab:ablation_ALL} is the adapted SGDet3D control with AGR, GCR,
% and DVC disabled. Controlled ablations use the same
% seed, schedule, and checkpoint-selection rule and are single runs; we report
% effect sizes rather than significance claims.

\subsection{Main Results}
\label{sec:main_results}

\begin{figure*}[t]
    \vspace{-10pt}
    \centering
    \begin{minipage}[t]{0.55\textwidth}
        \centering
        \includegraphics[width=\linewidth]{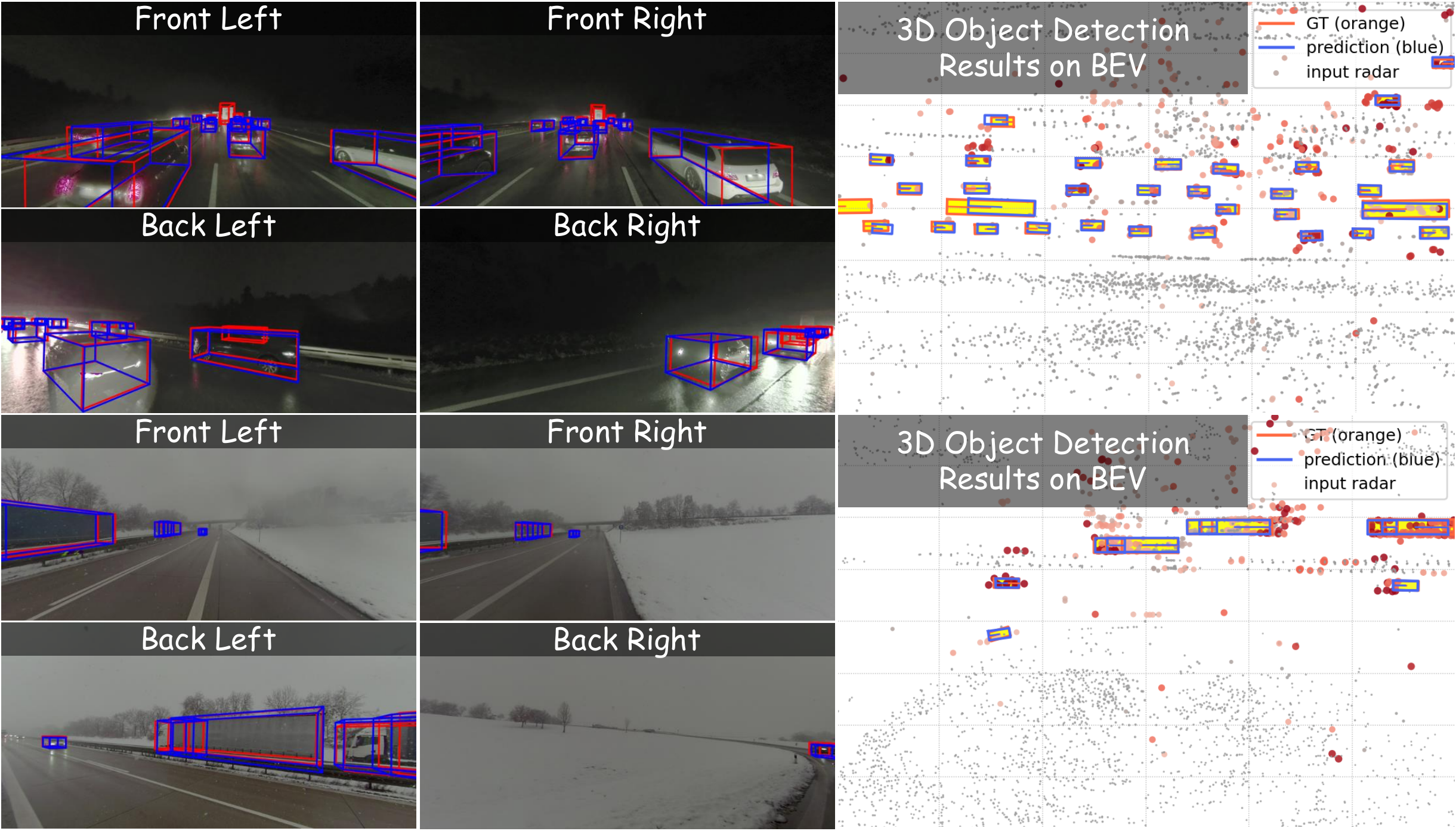}
    \end{minipage}
    \hfill
    \begin{minipage}[t]{0.435\textwidth}
        \centering
        \includegraphics[width=\linewidth]{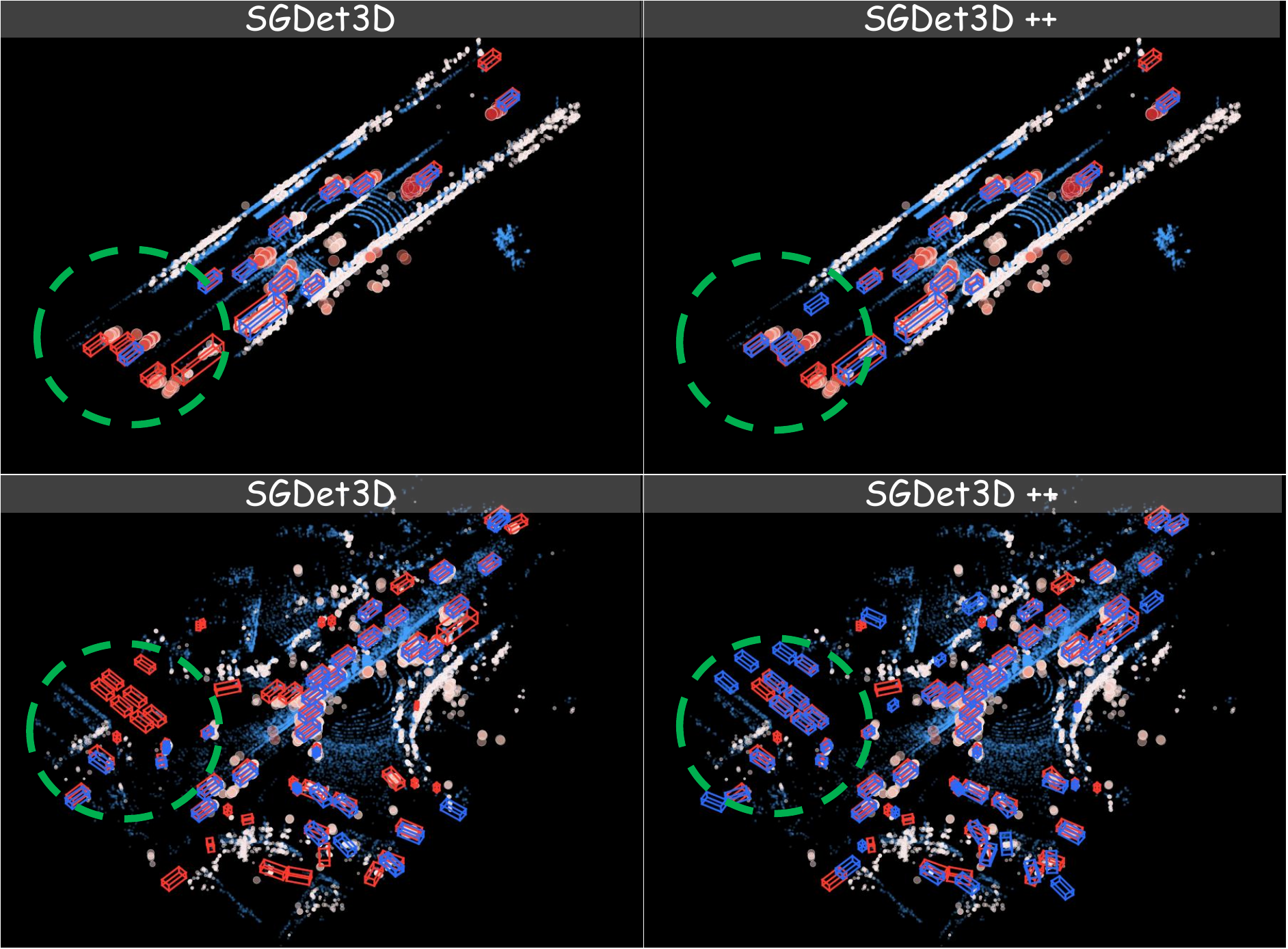}
    \end{minipage}
    \vspace{-8pt}
    \caption{Qualitative evidence on ManTruckScenes. \textit{Left:}
        \sgdetpp{} aligns multi-view detections with radar-supported BEV
        predictions. \textit{Right:} paired \sgdet{} and \sgdetpp{} outputs;
        green circles mark recovered detections and consolidated object boxes.
        Red boxes denote ground truth and blue boxes denote predictions.}
    \label{fig:mantruck_qualitative}
    \vspace{-10pt}
\end{figure*}

\begin{wraptable}[12]{r}{0.57\columnwidth}
    \vspace{-20pt}
    \belowrulesep=0pt
    \aboverulesep=0pt
    \caption{Adverse-condition on OmniHD-Scenes.}
    \label{tab_adverse_main}
    \centering
    \small
    \setlength{\tabcolsep}{1.4pt}
    \renewcommand\arraystretch{1.00}
    \begin{tabular}{@{}L{4.90cm}|C{0.90cm}|C{0.90cm}C{0.90cm}@{}}
        \toprule
        \makebox[1.0cm]{Method} & \makebox[0.90cm]{Mod.} &
        \makebox[0.90cm]{mAP} & \makebox[0.90cm]{ODS} \\
        \midrule
        PointPillars \citep{PointPillars} & R & 28.63 & 40.70 \\
        RadarPillarNet \citep{RCFusion} & R & 29.78 & 41.00 \\
        \midrule
        LSS \citep{LSS} & C & 21.21 & 25.62 \\
        % BEVFormer-S \citep{BEVFormer} & C & 28.11 & 29.36 \\
        BEVFormer \citep{BEVFormer} & C & 30.39 & 31.61 \\
        PanoOcc \citep{panoocc} & C & 26.09 & 27.16 \\
        % Sparse4Dv3 \citep{Sparse4Dv3} & C & 26.54 & 32.06 \\
        \hline
        BEVFusion \citep{BEVFusion} & R+C & 35.83 & 44.95 \\
        % M-CONet (ICCV'23) \citep{MCONet2023} & R+C & \na & \na \\
        RCBEVDet \citep{RCBEVDet} & R+C & 37.49 & 47.32 \\
        % Doracamom-S \citep{Doracamom2026} & R+C & 38.75 & 43.47 \\
        Doracamom \citep{Doracamom2026} & R+C & 41.86 & 48.74 \\
        4DR360$^\circ$ \citep{DR3602026} & R+C & \secondcell{44.18} & \secondcell{51.57} \\
        \rowcolor{gray!20}
        \sgdetpp{} & R+C & \bestcell{49.95} & \bestcell{58.86} \\
        \bottomrule
    \end{tabular}
    \vspace{-10pt}
\end{wraptable}
Tables~\ref{tab_omnihd_det} and \ref{tab_mantruck_det} compare 3D detection on
both benchmarks, where \sgdetpp{} performs best among the listed methods. On
OmniHD-Scenes, it raises the strongest R+C baseline \citep{DR3602026} from
45.05 to 48.87 mAP and from 51.40 to 58.22 ODS, improves all four classes, and
reduces mATE, mAOE, and mAVE. On ManTruckScenes, it raises the strongest listed
baseline from 49.57 to 56.39 mAP and from 52.97 to 62.19 NDS, with the largest
class gain on obstacles. These semantic, localization, and motion gains match
the complementary roles of AGR, GCR, and DVC. Figures~\ref{fig:qualitative_results}
and \ref{fig:mantruck_qualitative} align predictions with BEV radar evidence;
the latter also shows fewer missed or fragmented detections than \sgdet{}.

\subsection{Robustness and Decision-Level Evidence}
\label{sec:robustness}

\paragraph{Adverse-condition detection.}
Table~\ref{tab_adverse_main} evaluates the adverse OmniHD-Scenes subset.
Radar-only methods remain competitive, radar--camera methods generally exceed
camera-only detectors, and \sgdetpp{} obtains the best mAP and ODS. AGR grounds
image retrieval in radar support, while GCR and DVC retain anchor-local
geometry and Doppler-validated memory. Under camera-extrinsic perturbations,
\sgdetpp{} also outperforms an independently trained \sgdet{} control and
degrades more slowly: at $\pm10^\circ$ and $\pm1.0$\,m, it retains
41.70 mAP and 52.76 ODS, exceeding Control by 10.67 and 12.56 points.
% When extrinsic noise shifts image projections, GCR and DVC continue to use
% anchor-local radar geometry and Doppler, while AGR reweights the perturbed
% samples with anchor-associated radar evidence to reduce dependence on any
% single image projection.
% training and radar calibration remain unchanged.
% adapted SGDet3D baseline.

\begin{wraptable}[6]{r}{0.54\columnwidth}
    \vspace{-15pt}
    \belowrulesep=0pt
    \aboverulesep=0pt
    \centering
    \caption{Mechanism-targeted evidence for the three grounding decisions. Values
        are percentages.}
    \label{tab:grounding_evidence}
    \footnotesize
    \renewcommand\arraystretch{1.00}
    \setlength{\tabcolsep}{1.4pt}
    \begin{tabular}{@{}L{1.05cm}L{3.00cm}C{1.00cm}C{1.00cm}C{1.00cm}@{}}
        \toprule[1.0pt]
        Module & Diagnostic & Base & Ground & $\Delta$ \\
        \midrule
        AGR & Heavy-occ. AP@1m $\uparrow$ & 10.33 & 11.89 & +1.56 \\
        Gate & Target-return purity $\uparrow$ & 29.95 & 58.87 & +28.92 \\
        DVC & Motion-conflict F1 $\uparrow$ & 0.62 & 10.61 & +9.99 \\
        \bottomrule[1.0pt]
    \end{tabular}
    \vspace{-8pt}
\end{wraptable}
\noindent\textbf{Mechanism-level evidence.}
Table~\ref{tab:grounding_evidence} reports targeted tests: AGR uses the complete
Base/Base+AGR heavy-occlusion comparison, while the gate and DVC use the frozen
43-scene split. Relative to radius-only selection, the yaw-aligned box gate
raises target-return purity from 29.95\% to 58.87\% and lowers foreign returns
from 7.15\% to 1.16\%. With conflicts comprising only 0.31\% of eligible
histories, DVC reaches 10.61\% F1 while preserving 96.11\% of consistent
histories and retaining 75.90\% conflict recall.
AGR shifts 8.68\% of retrieval probability and changes the maximum-weight
camera--level--support route for 15.77\% of feature groups.

\subsection{Ablation Study}
\label{sec:ablation}

\paragraph{Component contribution.}
\begin{wraptable}[29]{r}{0.51\columnwidth}
    \vspace{-10pt}
    \centering
    \belowrulesep=0pt
    \aboverulesep=0pt
    \caption{Overall ablations on OmniHD-Scenes.}
    % Reported as mean$\pm$ standard deviation over three seeds.
    \label{tab:ablation_ALL}
    \footnotesize
    \setlength{\tabcolsep}{4.2pt}
    \renewcommand\arraystretch{1.00}
    \begin{tabular}{@{}C{0.80cm}|C{0.50cm}|C{0.50cm}|C{0.50cm}|C{1.63cm}|C{1.63cm}@{}}
        \toprule[1.0pt]
        \makebox[0.50cm]{\multirow{2}{*}{Base}} &
        \multicolumn{3}{c|}{Overall Ablation} &
        \makebox[0.80cm]{\multirow{2}{*}{mAP$\uparrow$}} &
        \makebox[0.80cm]{\multirow{2}{*}{ODS$\uparrow$}} \\
        \cmidrule(lr){2-4}
        & \makebox[0.50cm]{AGR} & \makebox[0.50cm]{GCR} & \makebox[0.50cm]{DVC} & & \\
        \midrule
        \checkmark & & & & \abresult{40.43}{0.28} & \abresult{47.82}{0.54} \\
        \checkmark & \checkmark & & & \abresult{43.94}{0.31} & \abresult{51.24}{0.27} \\
        \checkmark & & \checkmark & & \abresult{44.56}{0.16} & \abresult{52.07}{0.22} \\
        \checkmark & & & \checkmark & \abresult{42.29}{0.35} & \abresult{49.74}{0.39} \\
        \checkmark & \checkmark & \checkmark & & \abresult{45.68}{0.23} & \abresult{54.59}{0.20}\\
        \checkmark & \checkmark & \checkmark & \checkmark &
        \abresult{48.87}{0.45} & \abresult{58.22}{0.19} \\
        \bottomrule[1.0pt]
    \end{tabular}
    \belowrulesep=0pt
    \aboverulesep=0pt
    \centering
    \footnotesize
    % We report class-averaged AP@1m; $N$ is the number of evaluated GT objects.
    \caption{Class-averaged AP@1m by projected occlusion on
        OmniHD-Scenes. $N$ is the GT count.}
    \label{tab:occlusion_stratified}
    \setlength{\tabcolsep}{0.0pt}
    \renewcommand\arraystretch{1.00}
    \begin{tabular}{@{}L{2.00cm}C{1.70cm}C{1.70cm}C{1.70cm}@{}}
        \toprule[1.0pt]
        \makebox[1.58cm][l]{Method} &
        \makebox[1.62cm][c]{\makecell{Light\\$<20\%$}} &
        \makebox[1.62cm][c]{\makecell{Moderate\\$20$--$50\%$}} &
        \makebox[1.62cm][c]{\makecell{Heavy\\$\geq 50\%$}} \\
        \midrule
        GT objects ($N$) & 43{,}246 & 10{,}926 & 40{,}172 \\
        \midrule
        Base & 41.15 & 15.67 & 10.33 \\
        Base w/ AGR & 45.97 & 18.96 & 11.89 \\
        \rowcolor{gray!20}
        Full & \textbf{51.24} & \textbf{23.15} & \textbf{16.74} \\
        \bottomrule[1.0pt]
    \end{tabular}
    \centering
    \footnotesize
    \belowrulesep=0pt
    \aboverulesep=0pt
    \caption{Accuracy and resource comparison.}
    \label{tab:efficiency_supp}
    \setlength{\tabcolsep}{0.5pt}
    \renewcommand\arraystretch{1.00}
    \begin{tabular}{@{}L{1.65cm}|C{0.80cm}C{0.80cm}C{0.89cm}C{0.94cm}|C{0.90cm}C{0.90cm}@{}}
        \toprule[1.0pt]
        \multirow{2}{*}{Method} &
        \multicolumn{4}{c|}{Resource} &
        \multicolumn{2}{c}{Metrics}\\
        \cline{2-7}
        & Mem & Lat. & Param. & \makebox[0.91cm]{GFLOPs} & mAP$\uparrow$ & NDS$\uparrow$\\
        \midrule
        BEVFusion & 3.7 & 217.4 & 41.5 & 782.2 & 30.97 & 40.65\\
        RCFusion & 3.8 & 185.2 & 36.3 & 684.8 & 31.61 & 40.57\\
        LXL & 4.6 & 192.3 & 41.5 & 782.2 & 34.70 & 43.04\\
        SGDet3D$^\dagger$ & 9.9 & 333.3 & 65.9 & 1187.1 & 41.37 & 47.50\\
        RCBEVDet & 5.1 & 153.8 & 86.7 & 1650.9 & 41.72 & 46.85\\
        Doracamom & 9.6 & 192.3 & 55.8 & 724.2 & 41.98 & 46.36\\
        HGSFusion & 8.2 & 196.1 & 44.7 & 976.7 & 43.90 & 48.23\\
        4DR360$^\circ$ & 15.4 & 181.8 & 140.1 & 1936.8 & 49.57 & 52.97\\
        \rowcolor{gray!20}\textbf{\method{}} & 4.2 & 66.2 & 60.7 & 306.3 & 56.39 & 62.19\\
        \bottomrule[1.0pt]
    \end{tabular}
    \vspace{-10pt}
\end{wraptable}
Table~\ref{tab:ablation_ALL} reports the overall ablation of \sgdetpp{}.
Every component improves mAP and ODS over Base, with GCR giving the largest
isolated gain. AGR and GCR together add 5.25 mAP, and DVC adds another 3.19
mAP. Within GCR, jointly using local XYZ, RCS, and Doppler reaches 48.87
mAP and 58.22 ODS, the best result among the evaluated cue combinations.

To isolate the box gate in Equation~\ref{eq:anchor_consistent_returns}, we
remove only the yaw-aligned 3D box-membership test while keeping the radius
threshold and all other settings fixed. The radius-only variant loses 1.89 mAP
and 2.08 strict AP@1m, showing that box membership provides information beyond
proximity. The default configuration reaches 48.87 mAP, 58.22 ODS,
42.28 strict AP@1m, and 0.3350 mAVE.

\noindent\textbf{Detection under projected occlusion.}
Table~\ref{tab:occlusion_stratified} tests AGR under projected image occlusion.
We stratify visible GT objects by the fraction of their multi-view projection
occluded by closer objects. With GCR and DVC disabled, AGR improves AP@1m in
every stratum, by 4.82, 3.29, and 1.56 points in the light, moderate, and heavy
bins, respectively. A fixed-capacity control isolates the effect of memory
verification. At the same 600-slot budget, DVC improves mAP
from 46.12 to 48.87 and reduces mAVE from 0.3820 to 0.3350, directly
attributing the gain to Doppler verification.
% Together, these controls test the formulation at three levels: association
% geometry, occlusion-sensitive semantic access, and motion-consistent temporal
% inheritance. Their agreement with the predicted behavior supports
% geometry-grounded semantics beyond aggregate benchmark gains.

\noindent\textbf{Efficiency analysis.}
On one RTX A6000, Table~\ref{tab:efficiency_supp} reports memory in GB, latency
in ms, and parameters in millions. \sgdetpp{} attains the best mAP and NDS with
66.2\,ms latency, 4.2\,GB memory, 60.7\,M parameters, and 306.3 GFLOPs; it also
has the lowest latency and GFLOPs among the compared methods.
% The gain therefore does not arise from brute-force scaling.
% The complete system consequently advances the
% accuracy--efficiency balance.
%  Within this scaffold, AGR retrieves image features at anchor-consistent
% supports, GCR verifies the retrieved evidence with local 4D radar returns before
% refinement, and DVC reuses history only when object motion remains

\vspace{-10pt}
\section{Conclusion}
\label{sec:conclusion}

In this work, we present \sgdetpp{}, an object-level framework that grounds
each query's semantic retrieval, geometric refinement, and temporal inheritance
in object-conditioned 4D radar evidence. At every stage, the current anchor
selects radar evidence for the same object hypothesis, turning radar from a
generic proximity cue into object-specific geometric and motion evidence. AGR
retrieves radar-supported image features, GCR refines query geometry with
anchor-local returns, and DVC retains only historical hypotheses consistent with
current Doppler. Anchor-conditioned evidence use therefore links semantic
discrimination, spatial correction, and
temporal continuity within one decoding path. Across two benchmarks,
\sgdetpp{} improves detection, localization, and motion over the strongest
 baselines. Robustness and resource analyses further show reliable and
efficient detection.
Future work will extend object-level grounding to vision--language--action
models.
\section*{AI Use Statement}
Generative AI tools were used for literature organization, reviewer-style
feedback, and language editing. The authors verified all citations, technical
descriptions, analyses, and revisions, and take responsibility for the final
content of this work.
\bibliography{iclr2027}
\bibliographystyle{iclr2027}

\end{document}